# Decomposing Physician Disagreement in HealthBench

Satya Borgohain[1] Roy Mariathas

**Abstract.** We decompose physician disagreement in the HealthBench medical AI evaluation dataset to understand where variance resides and what observable features can explain it. Rubric identity accounts for 15.8% of met/not-met label variance but only 3.6–6.9% of disagreement variance; physician identity accounts for just 2.4%. The dominant 81.8% case-level residual is not reduced by HealthBench's metadata labels ($z = -0.22$, $p = 0.83$), normative rubric language (pseudo $R^2 = 1.2\%$), medical specialty (0/300 Tukey pairs significant), surface-feature triage (AUC $= 0.58$), or embeddings (AUC $= 0.485$). Disagreement follows an inverted-U with completion quality (AUC $=$ 0.689), confirming physicians agree on clearly good or bad outputs but split on borderline cases. Physician-validated uncertainty categories reveal that *reducible* uncertainty (missing context, ambiguous phrasing) more than doubles disagreement odds (OR $=$ 2.55, $p < 10^{-24}$), while *irreducible* uncertainty (genuine medical ambiguity) has no effect (OR $= 1.01$, $p = 0.90$), though even the former explains only $\sim 3\%$ of total variance. The agreement ceiling in medical AI evaluation is thus largely structural, but the reducible/irreducible dissociation suggests that closing information gaps in evaluation scenarios could lower disagreement where inherent clinical ambiguity does not, pointing toward actionable evaluation design improvements.[2]

## 1. Introduction

Large language models (LLMs) are rapidly becoming a primary source of medical information. As of early 2026, over 40 million people use ChatGPT daily for health questions, and more than 5% of all ChatGPT messages globally are health-related [1]. Among more than 1000 physicians surveyed, 45% use OpenEvidence, 15.6% use ChatGPT, 5% use Abridge, and 3% use Claude for clinical queries [2]. The scale of this adoption makes the reliability of medical AI evaluation increasingly consequential.

Evaluating LLM responses for healthcare-related questions requires expert human judgment, yet physicians frequently disagree on whether a model's response meets clinical standards. HealthBench [3], a large-scale medical AI evaluation dataset, exemplifies this challenge: 22.5% of cases produce physician disagreement. This phenomenon is well-documented across clinical domains: pathologists interpreting breast biopsies achieve only 75.3% concordance, with agreement dropping to 48% for borderline atypia cases [4]; DSM-5 field trials [5] found reliability ranging from very good ($\kappa = 0.60$–$0.79$) to unacceptable ($\kappa < 0.20$) across 23 diagnoses, with major depression falling in the "questionable" range ($\kappa = 0.28$) despite structured diagnostic criteria; and radiological disagreement rates have held steady at approximately 30–40% across specialties and response criteria [6]. Understanding what drives this disagreement in evaluations matters for anyone building protocols that depend on expert labels.

[1]Correspondence to: satya.borg@gmail.com

[2]Code can be found here: https://github.com/satyaborg/healthbench-physician-disagreement

Physician disagreement sets a structural ceiling on performance. HealthBench reports a macro $F_1 = 0.709$ for GPT-4.1 as the grader, with the model agreeing with physicians roughly as well as physicians agree with each other. Arora et al. attribute disagreement to "ambiguity in criteria, ambiguity in conversations and responses to be graded, and differences in clinical specialization, risk tolerance, perceived severity, communication style, and interpretation of instructions" [3]. The meta-evaluation shows physician-physician and model-physician agreement both ranging from 55% to 75%, and describes disagreement as "inherent in the exercise," driven by physicians' diverging views on what constitutes a *good* model response [3]. This characterization motivates a natural follow-up: what are the relative contributions of these sources? Mutisya et al. [7] suggest that evaluation designs relying on expert opinion rather than version-controlled clinical practice guidelines may amplify this disagreement. Similarly, Singhal et al. [8] show physician preference for LLM answers over clinician answers on most evaluation axes, while Bedi et al. [9] report LLM evaluations matching or exceeding clinician-clinician reliability (LLM-jury ICC = 0.47 vs. clinician-clinician ICC = 0.43), but the *sources* of the human disagreement ceiling have not been decomposed quantitatively. Zheng et al. [10] established that LLM judges achieve $> 80\%$ agreement with human evaluators, the same level as human-human agreement, raising the question of whether the agreement ceiling is an inherent property of the evaluation task rather than a limitation of any particular judge.

Kahneman et al. [11] provide a theoretical framework for understanding this variability. They decompose "system noise" in professional judgment into *level noise* (consistent individual differences) and *pattern noise* (case-specific variability in how individuals respond). A recent scoping review confirms that noise is a significant driver of practice variation in medical decision-making, with pattern noise documented in 7 of 14 studies [12]. Our analysis can be understood as a noise audit [11] of HealthBench: physician ICC (2.4%) corresponds to level noise, while the 81.8% case-level residual corresponds to pattern noise plus occasion noise.

The annotation literature has increasingly recognized that human label variation is not merely noise but a property of annotation tasks themselves [13], [14]. Aroyo and Welty [15] challenge the single-truth assumption in annotation, showing through seven "myths" that disagreement arises from ambiguity in content, underspecification in guidelines, and annotator differences - factors that map to our case-level residual, rubric effects, and physician effects, respectively. Basile et al. [16] argue that evaluation should model disagreement rather than force consensus, and Uma et al. [17] provide a comprehensive survey of methods for learning from annotator disagreement. This literature has focused primarily on subjective NLP tasks (toxicity, sentiment), and the question of how these variance sources compare in magnitude within structured rubric-based medical evaluation remains open.

HealthBench additionally provides a large-scale meta-evaluation dataset: 60,896 individual physician judgments, each a tuple of (rubric criterion, conversation, response, physician grade) across 29,511 unique cases, graded by 186 anonymized physicians (out of 262 total) with binary met/not-met labels. The dataset spans 34 unique rubric criteria, referred to as *consensus criteria* (30 unique rubric texts; 1 triple and 2 pairs share identical wording), with metadata on themes, categories, and evaluation axes. All variance decomposition models use the 34 consensus criteria as the grouping variable. With a median of 2 graders per case (range 2–5) and an overall disagreement rate of 22.5%, it provides a large sample size to decompose disagreement by source.

We ask: *Where does disagreement variance live, and can any observable feature explain it?* Our approach proceeds in nine phases: (1) label-level variance decomposition, (2) disagreement-level variance decomposition, (3) physician- and domain-level null results, (4) specialty contestedness ranking, (5) rubric language effects, (6) HealthBench metadata variance testing, (7) quality boundary effects, (8) predictive modeling with surface features and embeddings, and (9) consensus-validated uncertainty categories.

We find that the vast majority of disagreement variance is case-specific. While several drivers are detectable, each explains only $\sim 3\%$ of disagreement variance. For instance, rubric identity explains 15.8% of label variance, and reducible uncertainty doubles disagreement odds. This echoes the foundational finding by Elstein et al. [18] of "case specificity" in clinical reasoning, where physician performance on one case does not predict performance on the next, now extended to the domain of medical AI evaluation. Norman et al. [19] quantified this phenomenon using 6,342 medical students, finding $\sim 80\%$ of error variance at the item level, strikingly consistent with our 81.8% case-level residual.

## 2. Data and Methods

### 2.1. Dataset

| Property | Value |
|---|---|
| Source | HealthBench meta-evaluation dataset |
| Cases | 29,511 (prompt × completion × rubric) |
| Observations | 60,896 (one per physician × case) |
| Physicians | 186 (anonymized IDs) |
| Unique rubric criteria | 34 |
| Overall disagreement rate | 22.5% |
| Overall pass rate | 77.0% |
| Graders per case | 2 (median), range 2–5 |

Table 1: Dataset summary statistics.

**Consensus subset.** Phase 9 additionally uses the HealthBench Consensus dataset (711 prompts, 8,526 cases after joining), which contains physician-validated `example_tags` classifying each prompt's uncertainty type and context sufficiency. These tags were assigned through a separate consensus process and are exogenous to the meta-evaluation labels.

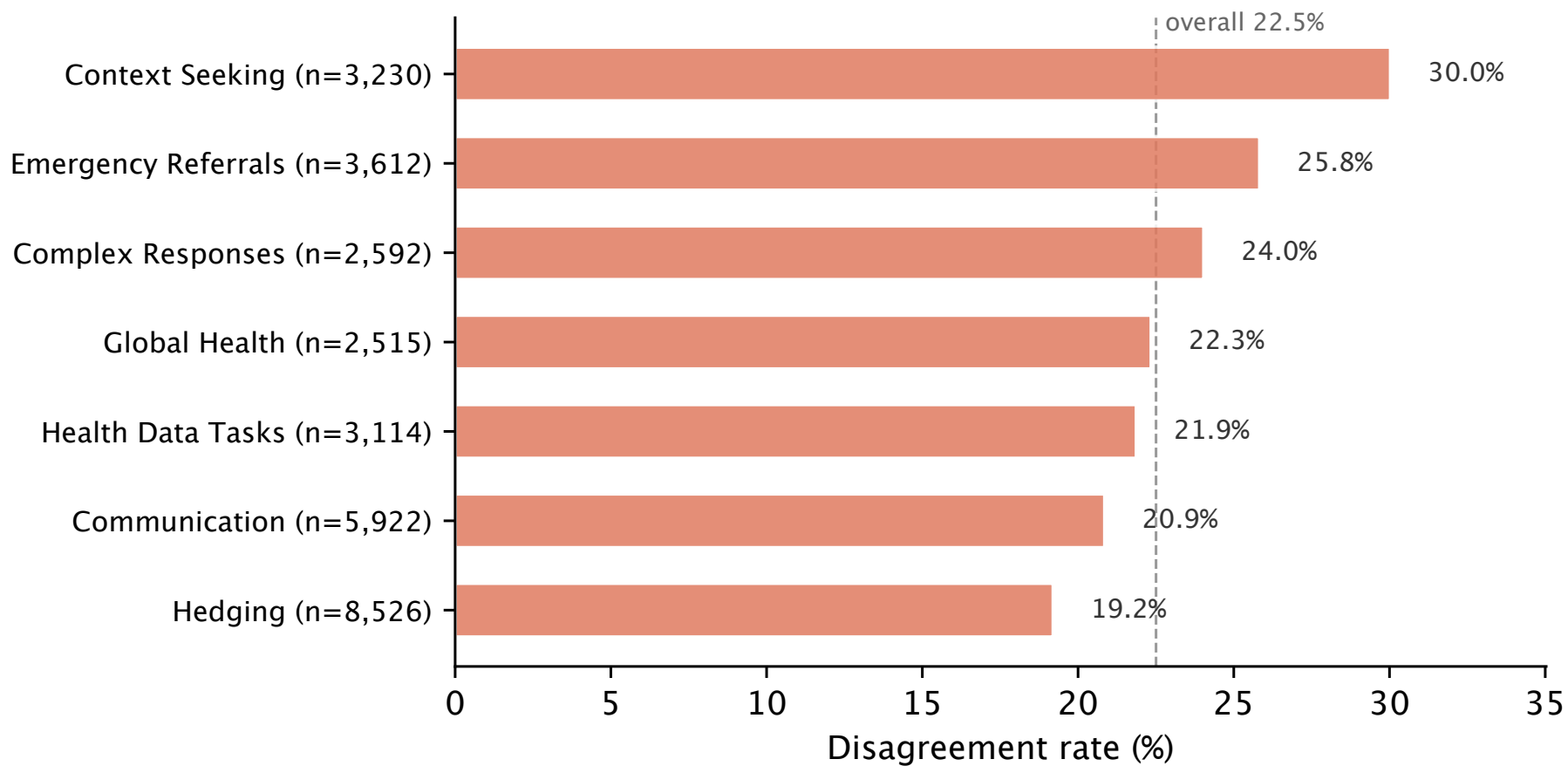

Figure 1: Disagreement rate by evaluation cluster. All seven clusters fall within a narrow 19–30% band. Context Seeking (30.0%) and Emergency Referrals (25.8%) show the highest disagreement; Hedging (19.2%) the lowest.

## 2.2. Modeling Approach

Primary variance decomposition uses linear mixed models (LMM) with physician random intercepts and rubric variance components, fitting a linear probability model (LPM) to binary outcomes. This decomposition follows the generalizability theory framework [20], [21], partitioning observed score variance into physician/rubric facets and their residual. ICCs are computed following [22], with form selection guided by [23]. Phase 2 adds a generalized linear mixed model (GLMM) robustness check with logit link to validate LPM ICCs on the latent scale. Logistic regressions use standard maximum likelihood estimation with clustered standard errors where observations are nested within prompts or criteria. Predictors are $z$-scored where noted. Mixed models are fit via REML for variance components and ML for likelihood ratio tests and information criteria, unless stated otherwise.

## 2.3. Classification Pipelines

**Specialty classification.** We use GPT OSS (120B) to assign primary and secondary medical specialties from 26 ABMS board categories per HealthBench's physician ontology [3]. Classification covers all 26 specialties with 95.9% domain match rate. Top specialties are Family Medicine (20.8%), Internal Medicine (20.5%), and Emergency Medicine (11.5%).

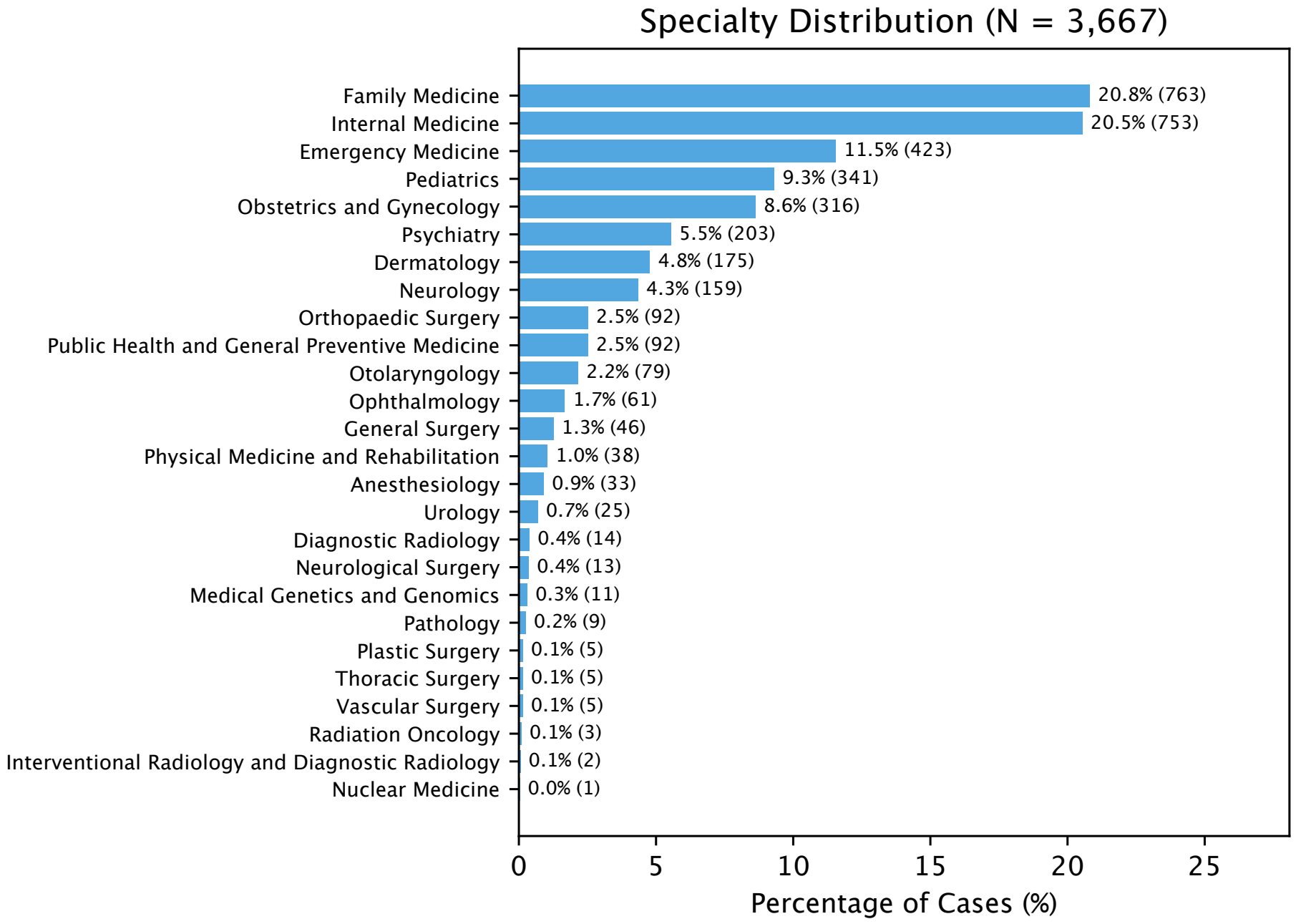

Figure 2: Distribution of LLM-assigned primary medical specialties across 3,667 HealthBench prompts with specialty assignments (of 3,671 total). Family Medicine and Internal Medicine together account for 41.3% of cases.

**Rubric feature extraction.** A 3-model ensemble (Claude Sonnet 4.5, GPT 5.2, Gemini 3 Pro) is used for classifying each rubric criterion as factual, procedural, or normative (Fleiss' $\kappa \approx 0.43$ [24]; moderate agreement per the [25] benchmark scale; $P_o = 0.80$, $P_e = 0.65$). Per-model normative ratios are: Claude 91.2%, GPT 76.3%, Gemini 70.6%, and a spread of 0.21 across models, which warrants caution when interpreting normative ratio effects.

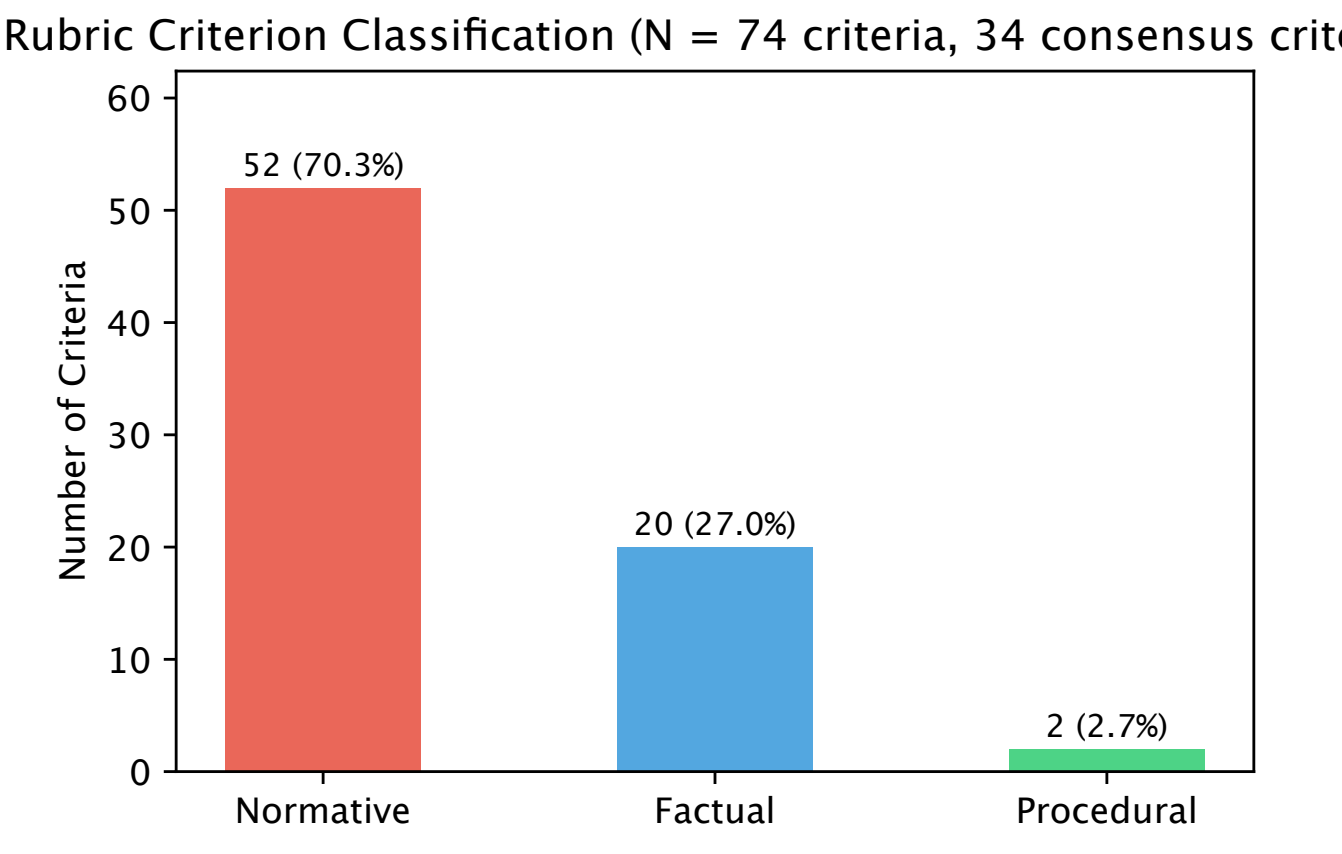

Figure 3: Distribution of rubric criterion types across 30 unique rubric texts (74 total criteria) by 3-model ensemble majority vote. 3 groups of consensus criteria (1 triple, 2 pairs) share identical rubric text. All variance decomposition models use the 34 consensus criteria as the grouping variable. Normative criteria (70.3%) dominate, consistent with the high normative ratios reported per model.

## 2.4. Disagreement Measures

We employ three case-level disagreement measures:

1. **Binary:** any split among raters = 1, unanimous = 0. Simple but treats a 1–1 split (2 raters) identically to a 2–3 split (5 raters) i.e., unweighted.
2. **Continuous strength:** $1 - |2 \cdot \text{pass_fraction} - 1|$ (0 = unanimous, 1 = perfect 50/50 split).
3. ***n*-weighted pairwise disagreement:**

$$D_{\text{pw}} = \frac{2n_{\text{pass}}n_{\text{fail}}}{n(n-1)} \tag{1}$$

The proportion of rater pairs that disagree. Properly weights by number of graders.

## 3. Results

### 3.1. Variance Decomposition (Label-Level)

For the first phase, we fit a linear mixed model: label $\sim 1 + (1 \mid \text{physician}) + (1 \mid \text{rubric})$

| Component | Variance | ICC |
|---|---|---|
| Physician | 0.0043 | 2.4% |
| Rubric | 0.0286 | 15.8% |
| Residual | 0.1477 | 81.8% |
| **Total** | **0.1807** | **100%** |

Table 2: Label-level variance decomposition. Rubric identity explains ≈ 7 times more variance than physician identity.

The LMM grand mean is 0.758, giving a theoretical Bernoulli variance of $p(1-p) = 0.18$. The observed residual (0.1477) falls below this ceiling, indicating that the random effects capture real structure rather than the residual being inflated by the binary outcome scale.

In the [11] noise taxonomy, the physician ICC of 2.4% corresponds to *level noise* (systematic individual differences in leniency), while the 81.8% residual encompasses *pattern noise* (case-by-physician interactions) and *occasion noise* (within-person stochasticity). The dominance of the residual is consistent with the general finding that pattern noise typically exceeds level noise in professional judgment [11], [12].

### 3.2. Variance Decomposition (Disagreement-Level)

To decompose variance in *disagreement* itself rather than met/not-met labels, we model case-level disagreement as the dependent variable with rubric as the sole random effect.

| Measure | Rubric ICC | Residual | Intercept |
|---|---|---|---|
| Binary (LPM) | 3.9% | 96.1% | 0.227 |
| Continuous strength (LPM) | 3.6% | 96.4% | 0.220 |
| $n$-weighted pairwise (LPM) | 3.6% | 96.4% | 0.220 |
| Binary (GLMM, latent scale) | 6.9% | 93.1% | — |

Table 3: Disagreement-level variance decomposition across measures and models. All measures point to rubric identity explaining very little disagreement variance.

The $n$-weighted measure (3.6%) closely matches continuous strength (3.6%), confirming that the binary disagreement ICC (3.9%) is not distorted by variable rater counts. The GLMM latent-scale ICC (6.9%) is modestly higher because the logit link maps binary outcomes to a continuous latent scale where the residual floor is $\pi^2/3 \approx 3.29$.

**Key contrast:** Rubric identity explains 15.8% of *whether a case passes* but only 3.6–6.9% of *whether physicians disagree*.

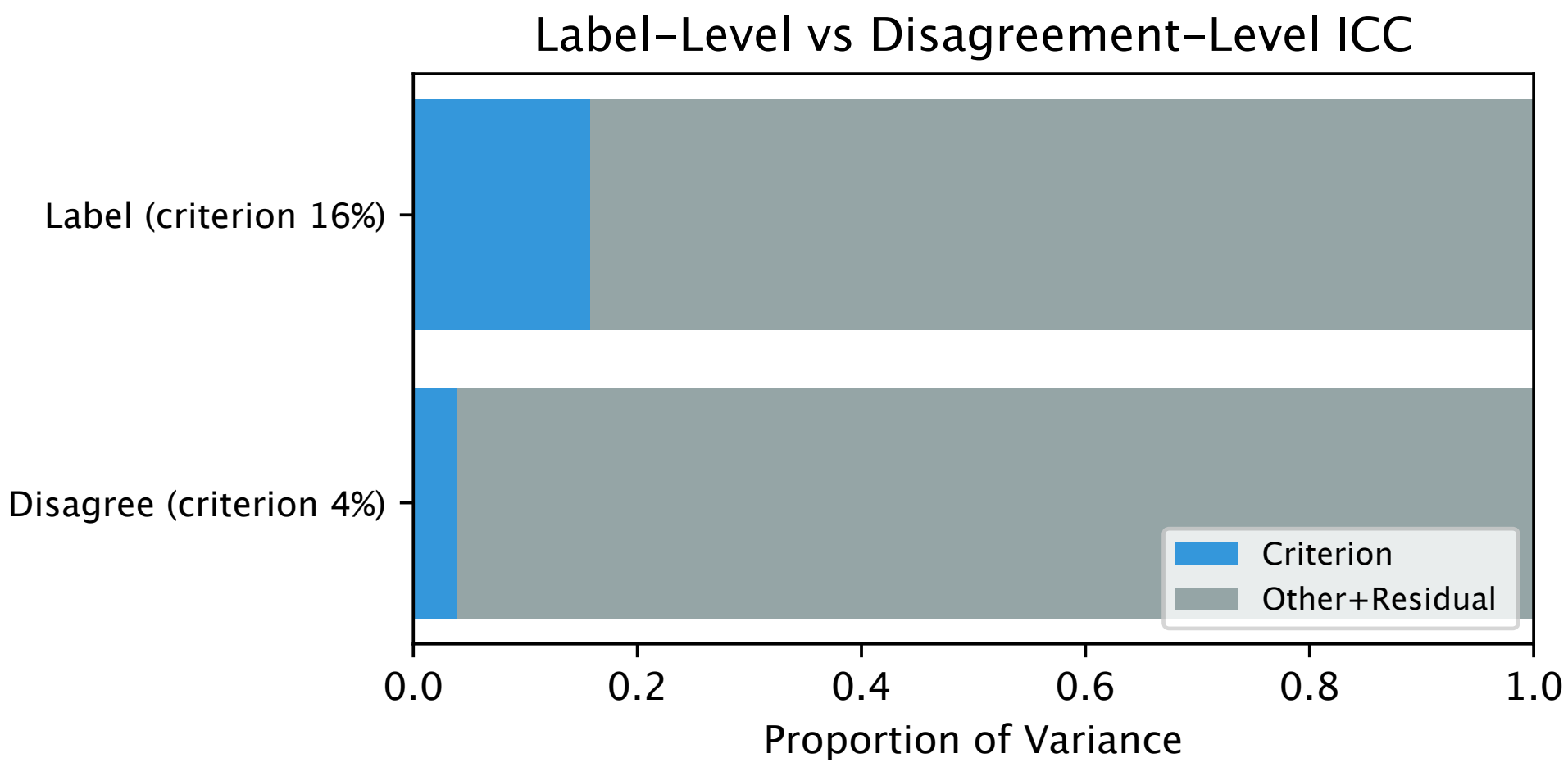

Figure 4: Label-level vs. disagreement-level variance decomposition. Rubric identity explains $\sim 16\%$ of met/not-met label variance but only $\sim 4\%$ of disagreement variance.

## 3.3. Physician and Domain-Level Effects

Two analyses test whether physician-level factors explain disagreement. **Domain match:** physician specialty is inferred via leave-one-out (LOO) assignment distributions (for each observation, the physician's modal specialty is computed excluding the focal case to prevent data leakage; mean concentration = 0.42). Logistic regression finds no effect of evaluating within one's specialty ($\beta = +0.012$, $p = 0.40$); however, the moderate proxy concentration makes this an underpowered test; a true domain match effect could exist but remain undetectable with this proxy. **Physician calibration:** seven outlier physicians (CV $> 0.30$, threshold $=$ mean $+$ 2SD) are identified, but removing their individual votes (not entire cases) and recomputing disagreement reduces the overall rate by only 0.06pp, consistent with Phase 1's 2.4% physician ICC (i.e., when individual physician biases are small relative to item-level variability, removing outlier votes has negligible effect on aggregate disagreement).

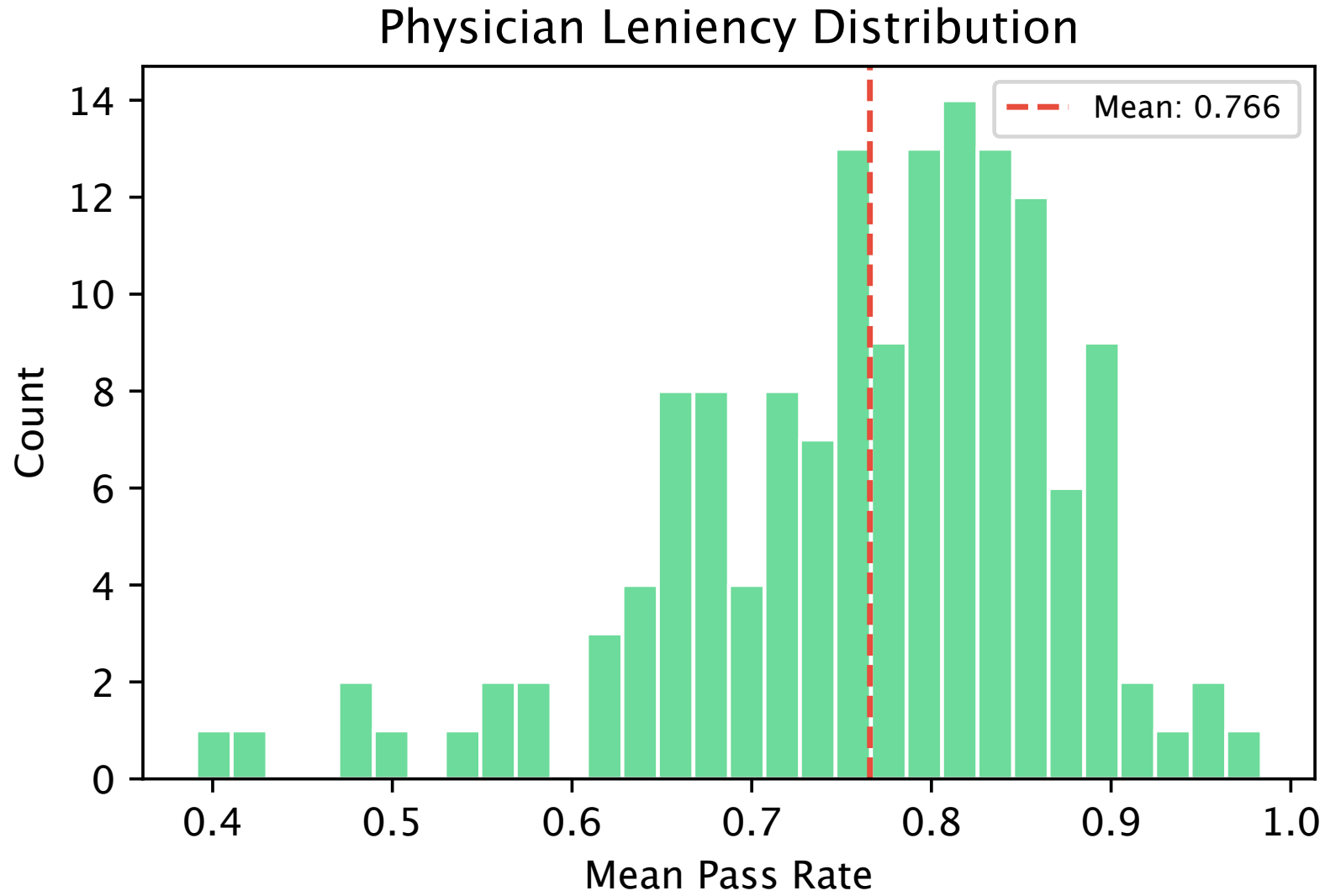

Figure 5: Distribution of physician-level pass rates (leniency). The tight clustering around the mean is consistent with the low physician ICC (2.4%); most physicians grade similarly.

### 3.4. Specialty Contestedness

ANOVA detects significant heterogeneity across 26 specialties ($F = 1.90$, $p = 0.005$), but zero pairwise comparisons survive Tukey HSD correction (0 of 300 pairs at $p < 0.05$). The effect reflects many small differences rather than a few outlier specialties, a pattern consistent with the diffuse variation documented across radiological specialties [6].

The highest-disagreement specialty (vascular surgery: 0.318, $n = 44$) has a small sample and wide CI. However, ophthalmology (0.250, $n = 508$) ranks second with a tight CI, suggesting a genuine elevation. Large-$N$ specialties cluster around the mean: family medicine $= 0.208$ ($n = 6,011$), emergency medicine $= 0.241$ ($n = 3,331$).

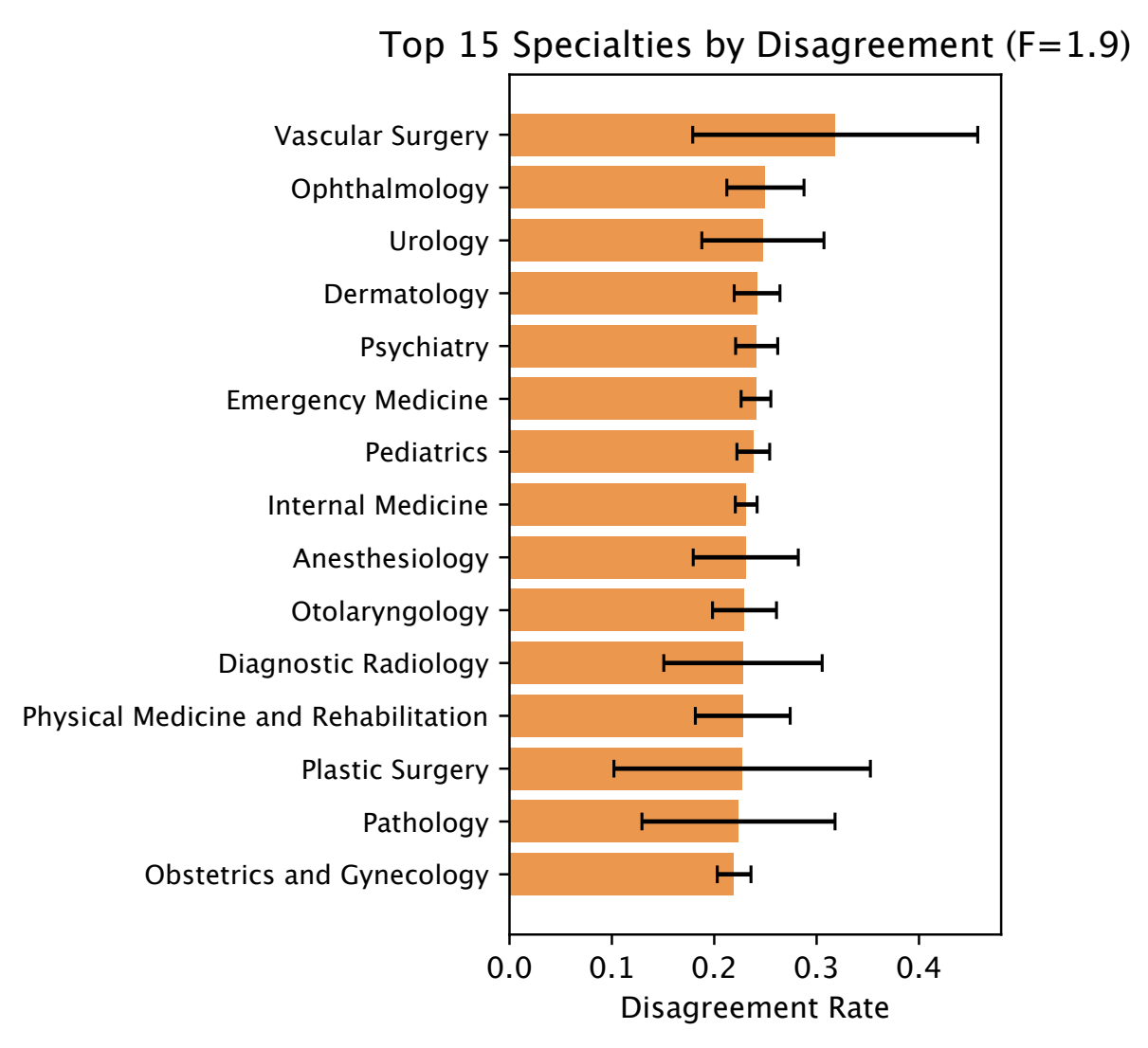

Figure 6: Top 15 medical specialties ranked by disagreement rate with 95% confidence intervals.

### 3.5. Rubric Language Effects

| **Predictor** | $\beta$ ($z$-scored) | $p$ |
|---|---|---|
| Normative ratio | +0.201 | 0.005 |
| Word count | +0.095 | 0.211 |
| Completion length (log) | +0.046 | 0.338 |

Table 4: Case-level logistic regression predicting disagreement ($N = 29,511$; clustered SEs on criterion, 34 clusters). Pseudo $R^2 = 0.012$.

With clustered standard errors on criterion (34 clusters), normative ratio is the only significant predictor ($p = 0.005$); word count ($p = 0.211$) and completion length ($p = 0.338$) lose significance once non-independence within criteria is accounted for. The pseudo $R^2$ of 1.2% means the effect is small in absolute terms. This is consistent with the rubric design literature: nonspecific evaluative criteria increase rater variability, but rubric specificity helps only at the margin and is unlikely to resolve case-level ambiguity on its own [26]. The rubric-level OLS ($N = 34$ consensus criteria) is significant ($R^2 = 0.19$, $p = 0.036$), confirming the case-level effect at the rubric level, though the Pearson correlation between normative ratio and disagreement rate is only marginally significant ($r = 0.31$, $p = 0.073$).

### 3.6. HealthBench Labels Variance Test

Next, we test whether HealthBench's metadata labels (theme, category, axis) absorb any of the 81.8% residual.

| **Model** | Phys% | Rubric% | Resid% | AIC |
|---|---|---|---|---|
| M0: baseline | 2.4% | 15.8% | 81.8% | 59,685 |
| M1: + theme | 1.8% | 14.3% | 83.8% | 59,501 |
| M2: + axis | 2.2% | 14.1% | 83.7% | 59,508 |
| M3: + all | 2.0% | 10.5% | 87.6% | 59,133 |

Table 5: Model comparison with HealthBench labels. Labels repartition rubric variance; residual is unchanged.

All models improve significantly over baseline by likelihood ratio test, and AIC favors M3. However, the improvement comes from repartitioning the rubric component (15.8% $\rightarrow$ 10.5%) rather than absorbing residual variance. The absolute residual variance changes by only $-0.00018$ ($-0.12\%$; $z = -0.22$, $p = 0.83$).

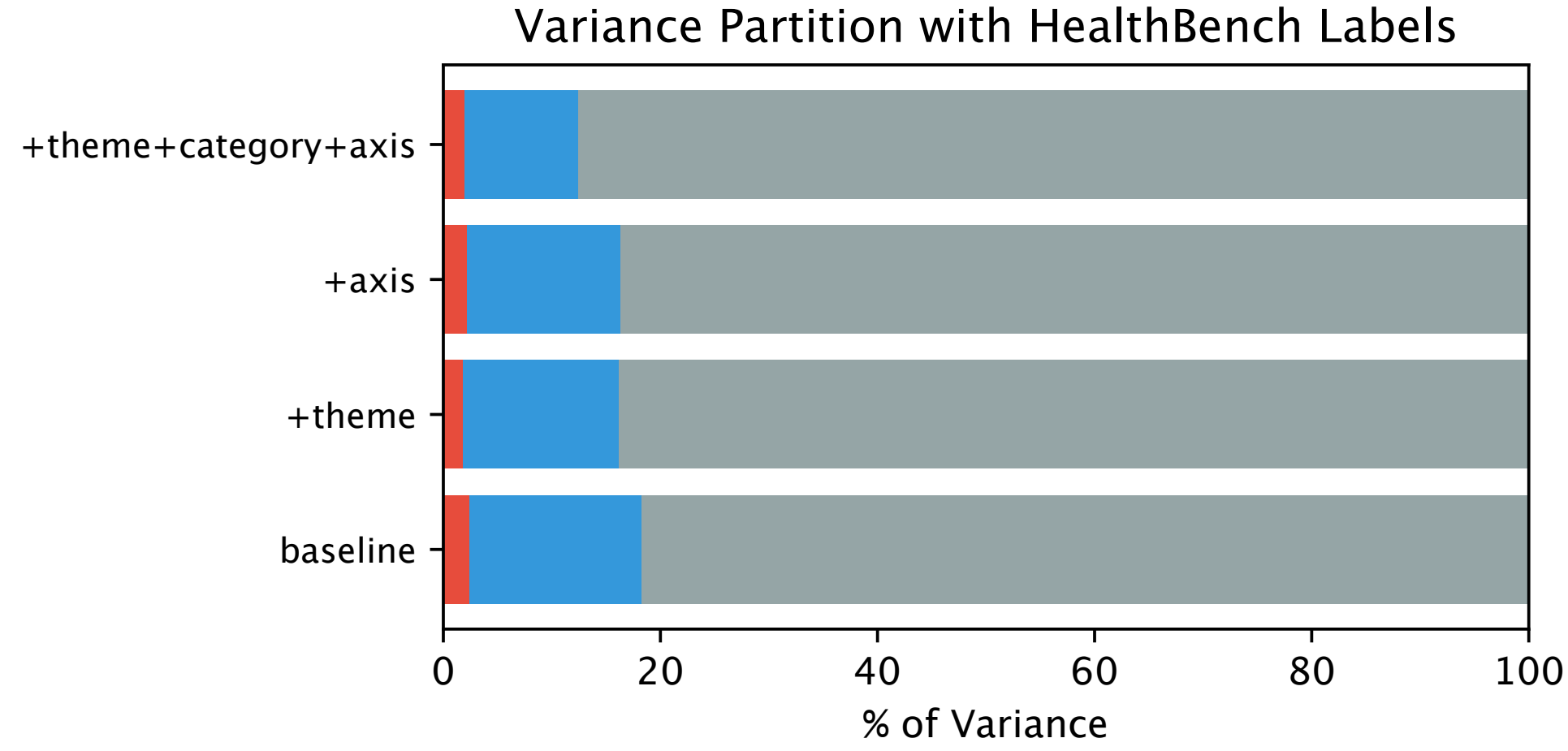

Figure 7: Variance partition across models with HealthBench labels. Adding metadata labels reduces rubric variance but leaves the residual unchanged.

### 3.7. Quality Boundary Effect

Disagreement is not uniformly distributed across completion quality. We use mean pass rate across physicians as a quality proxy and fit a quadratic logistic regression: $\text{disagree} \sim \text{quality} + \text{quality}^2$. Because the quality proxy is computed from the same physician labels that define disagreement, this analysis characterizes the *geometry* of disagreement in label space rather than identifying an independent predictor.

| **Metric** | **Value** | ***p*** |
|---|---|---|
| Quadratic term ($\beta_2$) | $-10.66$ | $< 10^{-124}$ |
| Pseudo $R^2$ | 8.3% | |
| AUC | 0.689 | |
| Leave-one-out pseudo $R^2$ | 3.0% | $< 10^{-24}$ |

Table 6: Quality–disagreement relationship (clustered SEs on prompt). The strongly significant negative quadratic term confirms an inverted-U relationship.

The inverted-U is clear, with disagreement at 38.5% in the lowest quality bin (pass rate $< 0.5$) but only 1.9% for near-unanimous cases (pass rate $> 0.94$). The leave-one-out analysis (pseudo $R^2 = 3.0\%$, still $p < 10^{-24}$) confirms a genuine effect beyond the mechanical floor imposed by the binary outcome structure: cases near 50% pass rate have more *room* for disagreement. Note: the leave-one-out quality proxy still shares the completion pool across cases within the same prompt, so some residual mechanical correlation may remain.

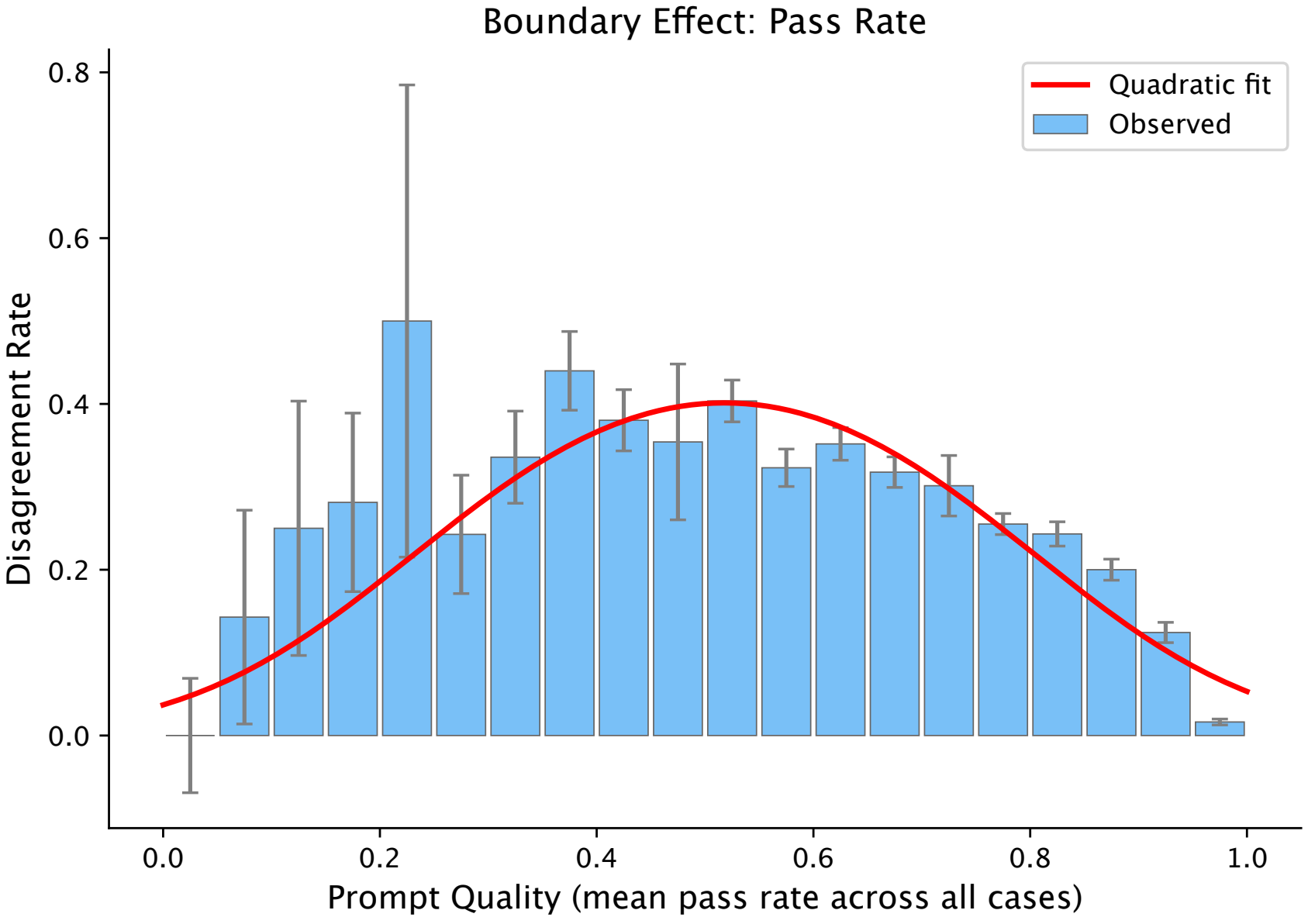

Figure 8: Disagreement rate as a function of completion quality (mean pass rate). The inverted-U confirms that physicians agree on clearly good or bad outputs but split on borderline cases.

## 3.8. Predictive Modeling (Surface Features and Embeddings)

**Surface-feature baseline.** A logistic regression model using word count, normative ratio, qualifier count, completion length, and grader count achieves 5-fold CV AUC $= 0.580 \pm 0.010$ (CV-aggregated AUC $= 0.579$), well above chance but far below practical utility. This contrasts with Schaekermann et al. [27], who designed ambiguity-aware AI systems that flag cases likely to produce expert disagreement in structured clinical tasks; in our setting, however, surface features provide insufficient signal for such flagging.

**Semantic embeddings.** Each case's full text (prompt, completion, rubric) is embedded via Google's `gemini-embedding-001` (3,072 dimensions).

| Model | AUC |
|---|---|
| PCA-50 logistic | 0.485 |
| Full embedding ($C = 0.1$) | 0.481 |
| Completion–rubric similarity | 0.488 |
| **Surface-feature baseline** | **0.580** |

Table 7: Disagreement prediction. Embedding models perform below the surface-feature baseline but they all have limited practical utility.

Centroid separation between agree and disagree cases yields cosine similarity $= 0.9998$; the two groups are geometrically indistinguishable. Within-group and between-group cosine similarities are identical ($\approx 0.809$).

Our best predictor (AUC 0.58) is consistent with the weak signal reported by Cheplygina and Pluim [28], whose crowd-disagreement features were similarly near-chance for melanoma classification, but remains below the AUC 0.78 reported by Raghu et al. [29], who achieved AUC 0.78 predicting expert disagreement in diabetic retinopathy, which was a structured grading task where image difficulty may provide a stronger signal. Taken together, these results are consistent with disagreement predictability varying with task structure: higher for standardized image grading, weak to absent for open-ended clinical QA evaluation, where the relevant distinctions may operate at a level of clinical reasoning granularity that surface and embedding features do not capture.

## 3.9. Consensus-Validated Uncertainty Categories

HealthBench's consensus dataset tags each prompt with an uncertainty category assigned through a separate physician consensus process: *any-reducible-uncertainty* (missing context, ambiguous phrasing), *only-irreducible-uncertainty* (genuine medical ambiguity), or *no-uncertainty.* These tags are exogenous to the meta-evaluation labels, providing a cleaner causal test than the LLM-inferred features used in earlier phases.

**Case-level logistic regression** ($N = 8,526$ cases, reference = no-uncertainty):

| Category | Disagree rate | OR vs. no-uncertainty | $p$ |
|---|---|---|---|
| No uncertainty ($n = 2,730$) | 13.2% | — | — |
| Irreducible only ($n = 2,376$) | 13.4% | 1.01 [0.82, 1.25] | 0.90 |
| Any reducible ($n = 3,420$) | 28.0% | 2.55 [2.13, 3.06] | $< 10^{-24}$ |

Table 8: Case-level disagreement by uncertainty category (clustered SEs on prompt). Reducible uncertainty more than doubles disagreement odds; irreducible uncertainty has no effect. Pseudo $R^2 = 3.4\%$, LLR $p < 10^{-60}$.

The dissociation is sharp: reducible uncertainty (28.0% disagreement) vs. irreducible uncertainty (13.4%) vs. no uncertainty (13.2%). Notably, physicians do not disagree more on cases with genuine medical ambiguity; disagreement is elevated when information is missing or the scenario is underspecified.

**Prompt-level robustness:** One-way ANOVA confirms the pattern ($F = 76.2$, $p = 1.1 \times 10^{-30}$). Chi-squared test on binary any-disagreement per prompt is also significant ($\chi^2 = 47.5$, $p = 4.8 \times 10^{-11}$).

**Context sufficiency:** A separate set of consensus tags classifies prompts as *enough-context* or *not-enough-context.* Among the 3,230 cases with context tags, insufficient context independently predicts higher disagreement (35.3% vs. 25.8%; OR = 1.56 [1.28, 1.91], $p = 1.3 \times 10^{-5}$; clustered SEs on prompt), corroborating the reducible-uncertainty finding.

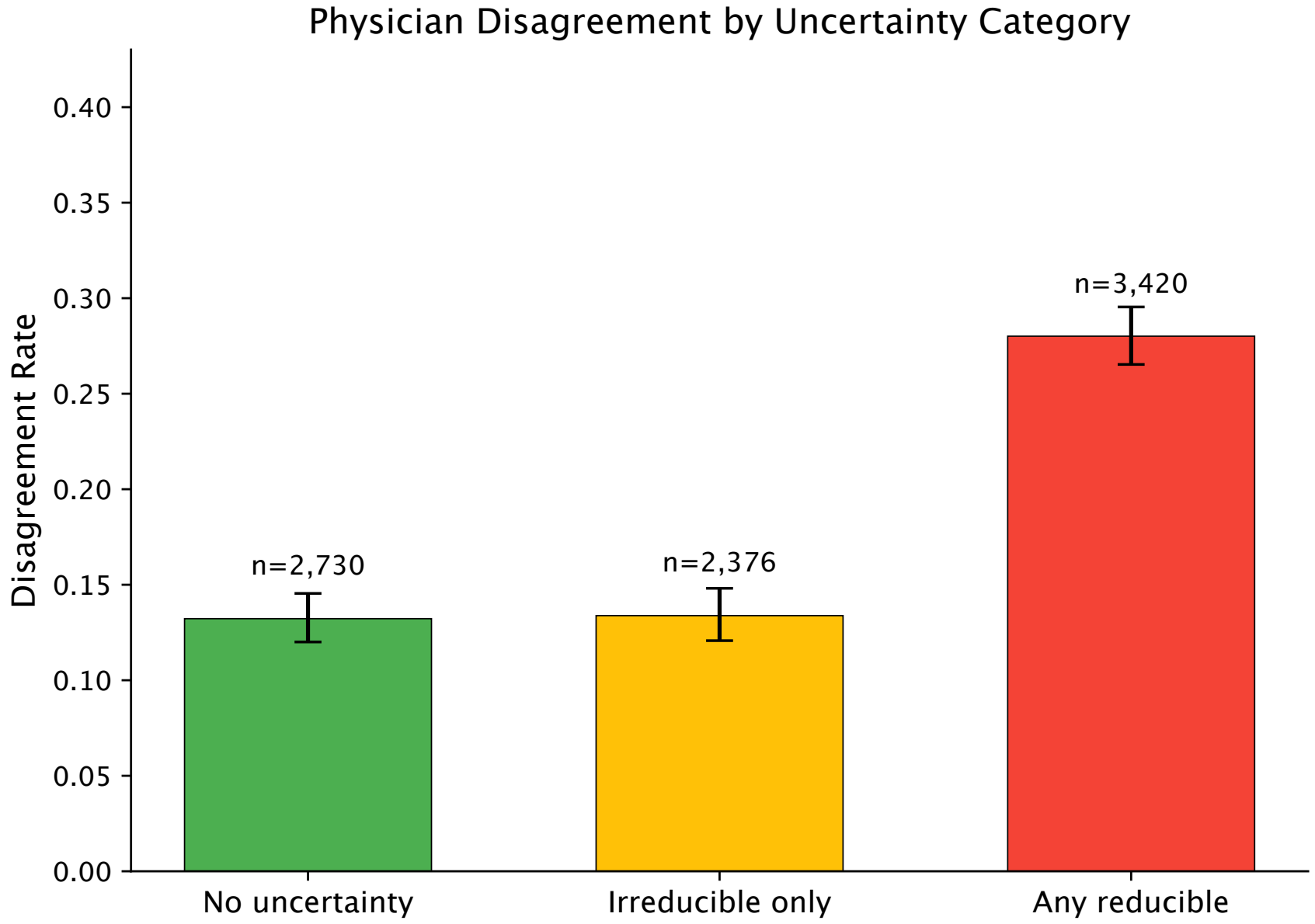

Figure 9: Disagreement rate by physician-validated uncertainty category with 95% CIs. Reducible uncertainty (28.0%) is sharply elevated relative to irreducible (13.4%) and no uncertainty (13.2%).

The pseudo $R^2$ of 3.4% is comparable to the quality boundary's leave-one-out $R^2$ (3.0%). The reducible/irreducible dissociation is theoretically important as it provides evidence against "inherent clinical ambiguity" as the primary driver of the case-level residual, and in favor of information gaps that are in principle addressable through better prompt and scenario design.

# 4. Discussion

## 4.1. Summary of Findings

Across all nine phases, the dominant pattern holds: physician disagreement in HealthBench is overwhelmingly case-specific. Two features, completion quality (Phase 7) and reducible uncertainty (Phase 9), show meaningful effects, each explaining $\sim 3\%$ of variance. The remaining $\sim 97\%$ remains unexplained by the features we examined.

| Phase | Question | Effect | $p$ | Interpretation |
|---|---|---|---|---|
| 1–2 | Where does variance live? | ICC: 2.4% / 15.8% / 81.8% | — | Case specificity dominates |
| 3 | Physician / domain effects? | $\beta = 0.012$; $\Delta = 0.06$pp | 0.40 | Individual differences small |
| 4 | Specialty differences? | $F = 1.90$ | 0.005 | Many small, no outlier specialties |
| 5 | Normative language? | $R^2 = 1.2\%$ | 0.005 | Detectable but practically small |
| 6 | Metadata absorbs residual? | $z = -0.22$ | 0.83 | Repartitions rubric, not residual |
| 7 | Quality boundary? | AUC = 0.689 | $< 10^{-124}$ | Inverted-U; partly mechanical |
| 8 | Surface features / embeddings? | AUC = 0.58 / 0.485 | — | Weak; semantic content not predictive |
| 9a | Reducible uncertainty? | OR = 2.55 | $< 10^{-24}$ | Information gaps double odds |
| 9b | Irreducible uncertainty? | OR = 1.01 | 0.90 | Genuine ambiguity has no effect |
| 9c | Context sufficiency? | OR = 1.56 | $< 10^{-5}$ | Corroborates reducible finding |

Table 9: Summary of all analysis phases.

## 4.2. What Drives the Residual?

None of the features we tested substantially reduce the 81.8% case-level residual (i.e., the sum of pattern noise and occasion noise per [11]). Phase 9's reducible/irreducible dissociation helps rank the candidate explanations:

**Information gaps** are the best-supported driver. Phase 9 shows that reducible uncertainty doubles disagreement odds (OR = 2.55), while irreducible uncertainty (genuine medical ambiguity) has no effect (OR = 1.01). However, the prompt-level tags explain only 3.4% of variance, because the relevant information gaps likely operate at the case level: within the same "reducible uncertainty" prompt, disagreement still varies across completions and rubrics. The prompt-level tag captures the *tendency* but not the case-specific interaction.

**Rubric-completion interactions** provide the structural explanation for why the residual is so large. In generalizability theory [20], [21], a dominant residual in a crossed design (physician $\times$ rubric $\times$ completion) reflects interactions that neither facet captures alone: a specific completion may be subtly ambiguous *with respect to* a specific rubric criterion in ways that neither completion features nor rubric features predict in isolation. This is consistent with the case specificity finding of [18] in clinical reasoning, now extended to rubric-based evaluation. Our Phase 8 null result (embeddings AUC = 0.485) corroborates this: the relevant distinctions are not in the semantic content of any single component but in their interaction.

**Ambiguity in the clinical scenario** is a natural hypothesis, but Phase 9 provides evidence *against* it as the primary driver. Prompts tagged as having only irreducible uncertainty show the same disagreement rate (13.4%) as those with no uncertainty (13.2%). Interestingly, genuine medical ambiguity, as identified by physician consensus, does not appear to elevate disagreement. Schaekermann et al. [30] distinguish "resolvable" from "irresolvable" disagreement in structured adjudication; the detectable portion of disagreement in HealthBench skews toward the resolvable type, though the small effect size ($R^2 = 3.4\%$) means most disagreement remains unexplained by either category.

**Stochastic physician judgment** may account for a substantial fraction of the residual as irreducible occasion noise [11]; the same physician shown the same case twice might grade differently. The Jackson et al. [31] finding of only 53% intra-observer agreement for atypia cases (95% CI 47–59%) suggests this contribution could be large, potentially 20–40% of the residual, but we have no test-retest data to bound it (see Limitations).

## 4.3. Implications for Medical AI Evaluation

HealthBench's macro $F_1 = 0.709$ for GPT-4.1 grader is likely best understood not as a model limitation (relative to human judgment) but as reflecting the $\approx 22.5\%$ case-level disagreement rate mechanically capping achievable $F_1$. This is partly because easy-consensus cases dominate the aggregate: interphysician macro $F_1$ ranges from 0.57 to 0.73 across themes. A complementary question worth exploring is how well automated graders perform specifically on cases where physicians themselves cannot agree. In our analysis, the 81.8% case-level residual is not substantially reduced by rubric design differences (3.6–6.9% of disagreement variance), specialty filtering (0/300 Tukey pairs significant), or metadata covariates ($z = -0.22$, $p = 0.83$). Phase 9 shows *some* of this ceiling is addressable through better prompt design (OR = 2.55 for reducible uncertainty), but the small effect size ($R^2 = 3.4\%$) implies most would remain. GPT-4.1 navigates the same irreducible ambiguity that physicians cannot resolve. Future benchmarks may benefit from contextualizing model-physician agreement against the variance structure of human labels, beyond aggregate agreement rates.

This has consequences for how benchmark results are interpreted more broadly. Chen et al. [32] show that when ground-truthing reproducibility is 70%, AI concordance can range from 40–100%, directly quantifying how label noise propagates into performance metrics. When physician labels are collapsed to a single "correct" answer, case-level uncertainty is effectively treated as error, which can conflate model mistakes with unresolved physician disagreement [33]. Preserving the full label distribution [13], [34] would let benchmarks distinguish "model got it wrong" from "model agreed with the minority physician." Kopanichuk et al. [35] propose metrics (RPAD/RRAD) that compare AI outputs against multiple expert opinions rather than a single reference, finding that expert variability often exceeds AI-human variability. The same concern extends to automated evaluation: LLM-as-judge systems validated against human labels inherit the case-level residual that we document. That LLM-jury ICC (0.47) exceeds clinician-clinician ICC (0.43) [9] likely reflects greater self-consistency (lower level noise) rather than superior judgment, since both are bounded by the same case-level ambiguity floor.

## 4.4. Limitations

The most significant constraint is the low rater count: 94.1% of cases have exactly 2 graders, so most disagreements are a 1–1 tie with no majority direction, preventing continuous disagreement measures and directional analysis for most cases. Our use of a linear probability model on binary data is a related simplification, though the GLMM robustness check yields a modestly higher latent-scale ICC (6.9% vs. 3.9%) that is qualitatively concordant.

On the measurement side, HealthBench defines 34 consensus criteria (30 unique texts), leaving rubric-level analyses underpowered, though the rubric-level OLS remains significant ($p = 0.036$). Several of our proxies potentially carry noise: physician specialty is inferred from assignment patterns (mean concentration $= 0.42$) rather than self-reported, and the rubric classification ensemble has only moderate inter-model agreement (Fleiss' $\kappa \approx 0.43$) [24], with a 20.6pp spread on normative classification suggesting ambiguity in what constitutes "normative" language.

Our embedding analysis uses only general-purpose models; domain-specific alternatives (e.g., BioLinkBERT) might capture finer reasoning-level distinctions. Similarly, fully crossed random effects [22] would be more appropriate than our specification but are computationally challenging at this scale; adding a physician×criterion interaction term splits the criterion component in half but leaves the 81.8% residual unchanged.

Finally, we cannot separate pattern noise from occasion noise because no physician evaluated the same case twice [11]; [31] demonstrates this is feasible in pathology. Although HealthBench rubrics carry point weights ($-10$ to $+10$) for aggregate scoring, the meta-evaluation elicits binary met/not-met labels per criterion [3], and our analysis operates on these native binary judgments. The binary format itself may limit observable variance structure: [9] reports clinician-clinician ICC = 0.43 on Likert scales, substantially higher than our 2.4%.

### 4.5. Future Directions

The central open question is how much of the 81.8% residual is even explainable. Our analysis cannot distinguish pattern noise (case-specific but systematic) from occasion noise (stochastic within-physician variability), and the answer determines whether richer features are worth pursuing at all. The highest-priority future direction is therefore physician self-consistency testing: presenting the same case to the same physician twice in different sessions would directly bound occasion noise [11], [31]. If occasion noise accounts for, say, 30–40% of the residual, as Jackson et al.'s [31] 53% intra-observer agreement for atypia suggests is plausible, then the ceiling on what any feature could explain drops accordingly.

The reducible/irreducible dissociation points to a second priority: case-level information gap annotation. The prompt-level consensus tags explain only 3.4% of variance partly because they operate at a different granularity - within the same "reducible uncertainty" prompt, disagreement still varies across completions and rubrics. Having physicians flag which specific rubric-completion pairs suffer from missing context or ambiguous phrasing would test whether the information gap finding scales when measured at the level where the variance actually lives. This follows directly from our finding that the residual is dominated by rubric-completion interactions rather than properties of any single component. Obtaining direct domain match labels via LLM assessment of (physician specialty × case) pairs would similarly replace the weak proxy that rendered domain match effects undetectable in our analysis.

Increasing rater counts per case would complement both directions: more raters improve reliability through averaging [20], [21] if the residual is occasion noise, or better characterize the disagreement distribution if it reflects genuine ambiguity. Increasing the rubric count beyond $N = 34$ would strengthen rubric-level inference. On the reporting side, disagreement-aware metrics that separate model predictions on consensus cases from contested cases could help distinguish genuine capability from alignment with one physician faction, complementing aggregate $F_1$ scores. Multi-annotator modeling frameworks [36], [37] remain a natural complement to this shift.

## 5. Conclusion

We decomposed physician disagreement in HealthBench across nine analysis phases and found that the vast majority of variance resides at the case level (81.8%). Rubric identity, physician identity, medical specialty, normative language, HealthBench metadata, surface features, and semantic embeddings each account for only a small share of this variance. Among the features we tested, two show the most promise: completion quality exhibits an inverted-U relationship with disagreement (AUC = 0.689; LOO $R^2 = 3.0\%$), and reducible uncertainty, as identified through physician-validated consensus tags, more than doubles disagreement odds (OR = 2.55; $R^2 = 3.4\%$). Most notably, irreducible uncertainty (genuine medical ambiguity as judged by physician consensus) shows no detectable association with disagreement (OR = 1.01), suggesting that information gaps and fine-grained reasoning interactions contribute more to the residual than inherent clinical ambiguity does. Taken together, these findings indicate that physician disagreement in medical AI evaluation largely reflects case-level pattern noise [11] that observable features do not readily capture. The reducible/irreducible dissociation does, however, point to one actionable lever: targeting information gaps at the case level is the most promising avenue for reducing addressable disagreement. Yet the modest effect sizes ($\sim 3\%$ $R^2$) suggest that much of the agreement ceiling is structural, underscoring the importance of accounting for this irreducible variance when interpreting benchmark results. Case-level information gap annotation and disagreement-aware evaluation metrics offer the most evidence-based paths forward, as better rubrics or automated triage alone will not move the ceiling.

## References

[1] OpenAI, "ChatGPT in Health." [Online]. Available: https://openai.com/index/introducing-chatgpt-health/

[2] OffCall, "AI Adoption Among Physicians: Survey of 1,000 Physicians." [Online]. Available: https://2025-physicians-ai-report.offcall.com/

[3] R. K. Arora and others, "HealthBench: Evaluating Large Language Models Towards Improved Human Health," *arXiv preprint arXiv:2505.08775*, 2025.

[4] J. G. Elmore, G. M. Longton, P. A. Carney, and others, "Diagnostic Concordance Among Pathologists Interpreting Breast Biopsy Specimens," *JAMA*, vol. 313, no. 11, pp. 1122–1132, 2015.

[5] D. A. Regier, W. E. Narrow, D. E. Clarke, and others, "DSM-5 Field Trials in the United States and Canada, Part II: Test-Retest Reliability of Selected Categorical Diagnoses," *American Journal of Psychiatry*, vol. 170, no. 1, pp. 59–70, 2013.

[6] A. M. Schmid *et al.*, "Radiologists and Clinical Trials: Part 1 The Truth About Reader Disagreements," *Therapeutic Innovation & Regulatory Science*, vol. 55, no. 6, pp. 1111–1121, 2021, doi: 10.1007/s43441-021-00316-6.

[7] F. Mutisya and others, "Rethinking Evidence Hierarchies in Medical Language Benchmarks: A Critical Evaluation of HealthBench," *arXiv preprint arXiv:2508.00081*, 2025.

[8] K. Singhal, T. Tu, J. Gottweis, and others, "Toward Expert-Level Medical Question Answering with Large Language Models," *Nature Medicine*, vol. 31, no. 3, pp. 943–950, 2025.

[9] S. Bedi, H. Cui, M. Fuentes, and others, "Holistic Evaluation of Large Language Models for Medical Tasks with MedHELM," *Nature Medicine*, 2026.

[10] L. Zheng, W.-L. Chiang, Y. Sheng, and others, "Judging LLM-as-a-Judge with MT-Bench and Chatbot Arena," in *NeurIPS 2023 Datasets and Benchmarks Track*, 2023.

[11] D. Kahneman, O. Sibony, and C. R. Sunstein, *Noise: A Flaw in Human Judgment*. Little, Brown Spark, 2021.

[12] K. V. Dlugos, M. Mazwi, R. Lao, and others, "Noise is an Underrecognized Problem in Medical Decision Making and is Known by Other Names: A Scoping Review," *BMC Medical Informatics and Decision Making*, vol. 25, p. 86, 2025.

[13] B. Plank, "The "Problem" of Human Label Variation: On Ground Truth in Data, Modeling and Evaluation," in *Proceedings of EMNLP*, 2022, pp. 10671–10682.

[14] A. Uma *et al.*, "SemEval-2021 Task 12: Learning with Disagreements," in *Proceedings of SemEval-2021*, 2021.

[15] L. Aroyo and C. Welty, "Truth Is a Lie: Crowd Truth and the Seven Myths of Human Annotation," *AI Magazine*, vol. 36, no. 1, pp. 15–24, 2015.

[16] V. Basile *et al.*, "We Need to Consider Disagreement in Evaluation," in *Proceedings of the 1st Workshop on Benchmarking: Past, Present and Future*, 2021, pp. 15–21.

[17] A. N. Uma, T. Fornaciari, D. Hovy, S. Paun, B. Plank, and M. Poesio, "Learning from Disagreement: A Survey," *Journal of Artificial Intelligence Research*, vol. 72, pp. 1385–1470, 2021.

[18] A. S. Elstein, L. S. Shulman, and S. A. Sprafka, *Medical Problem Solving: An Analysis of Clinical Reasoning*. Harvard University Press, 1978.

[19] G. Norman, G. Bordage, G. Page, and D. Keane, "How Specific Is Case Specificity?," *Medical Education*, vol. 40, no. 7, pp. 618–623, 2006.

[20] L. J. Cronbach, G. C. Gleser, H. Nanda, and N. Rajaratnam, *The Dependability of Behavioral Measurements: Theory of Generalizability for Scores and Profiles*. Wiley, 1972.

[21] R. L. Brennan, *Generalizability Theory*. Springer, 2001.

[22] P. E. Shrout and J. L. Fleiss, "Intraclass Correlations: Uses in Assessing Rater Reliability," *Psychological Bulletin*, vol. 86, no. 2, pp. 420–428, 1979.

[23] T. K. Koo and M. Y. Li, "A Guideline of Selecting and Reporting Intraclass Correlation Coefficients for Reliability Research," *Journal of Chiropractic Medicine*, vol. 15, no. 2, pp. 155–163, 2016.

[24] J. L. Fleiss, "Measuring Nominal Scale Agreement Among Many Raters," *Psychological Bulletin*, vol. 76, no. 5, pp. 378–382, 1971.

[25] J. R. Landis and G. G. Koch, “The Measurement of Observer Agreement for Categorical Data,” *Biometrics*, vol. 33, no. 1, pp. 159–174, 1977.

[26] J. J. Isaacson and A. S. Stacy, “Rubrics for Clinical Evaluation: Objectifying the Subjective Experience,” *Nurse Education in Practice*, vol. 9, no. 2, pp. 134–140, 2009.

[27] M. Schaekermann, G. Beaton, E. Sanoubari, A. Lim, K. Larson, and E. Law, “Ambiguity-Aware AI Assistants for Medical Data Analysis,” in *Proceedings of CHI 2020*, 2020.

[28] V. Cheplygina and J. P. W. Pluim, “Crowd Disagreement About Medical Images Is Informative,” in *LABELS 2018, Springer LNCS 11043*, 2019, pp. 105–111.

[29] M. Raghu *et al.*, “Direct Uncertainty Prediction for Medical Second Opinions,” in *Proceedings of ICML 2019*, 2019, pp. 5281–5290.

[30] M. Schaekermann, G. Beaton, M. Habib, A. Lim, K. Larson, and E. Law, “Understanding Expert Disagreement in Medical Data Analysis Through Structured Adjudication,” *Proceedings of the ACM on Human-Computer Interaction*, vol. 3, no. CSCW, p. Article142, 2019.

[31] S. L. Jackson *et al.*, “Diagnostic Reproducibility: What Happens When the Same Pathologist Interprets the Same Breast Biopsy Specimen at Two Points in Time?,” *Annals of Surgical Oncology*, vol. 24, no. 5, pp. 1234–1241, 2017.

[32] P.-H. C. Chen, C. H. Mermel, and Y. Liu, “Evaluation of Artificial Intelligence on a Reference Standard Based on Subjective Interpretation,” *The Lancet Digital Health*, vol. 3, no. 11, pp. e693–e695, 2021, doi: 10.1016/S2589-7500(21)00216-8.

[33] A. Sylolypavan and others, “The Impact of Inconsistent Human Annotations on AI-Driven Clinical Decision Making,” *npj Digital Medicine*, vol. 6, p. 26, 2023.

[34] F. Cabitza, A. Campagner, and V. Basile, “Toward a Perspectivist Turn in Ground Truthing for Predictive Computing,” in *Proceedings of AAAI 2023*, 2023, pp. 6860–6868.

[35] I. Kopanichuk and others, “How to Evaluate Medical AI,” *arXiv preprint arXiv:2509.11941*, 2025.

[36] A. M. Davani, M. Diaz, and V. Prabhakaran, “Dealing with Disagreements: Looking Beyond the Majority Vote in Subjective Annotations,” *Transactions of the Association for Computational Linguistics*, vol. 10, pp. 92–110, 2022.

[37] M. L. Gordon *et al.*, “Jury Learning: Integrating Dissenting Voices into Machine Learning Models,” in *Proceedings of CHI '22*, 2022.